\def\year{2019}\relax
\pdfoutput=1
\documentclass[letterpaper]{article} 
\usepackage{aaai19}  
\usepackage{times}  
\usepackage{helvet}  
\usepackage{courier}  
\usepackage{url}  
\usepackage{graphicx}  
\usepackage{graphicx, amsmath, subcaption, amssymb, comment, bm, amsthm, bbm}
\DeclareMathOperator*{\argmax}{arg\,max}

\usepackage{qtree}
\newcommand{\BR}{\mathtt{BR}}
\newcommand{\REP}{\mathtt{REPLACE}}
\newcommand{\JP}{\mathtt{JP}}
\newcommand{\CTH}{\mathtt{CTH}}
\newcommand{\tuple}[1]{\langle #1\rangle}
\newcommand{\citet}[1]
  {\citeauthor{#1} ~\shortcite{#1}}

\title{Theory of Minds: Understanding Behavior in Groups Through Inverse Planning}
\author{
  Michael Shum$^*$ \\ Brain and Cognitive Sciences \\ MIT \\ \ mshum@mit.edu
  \And Max Kleiman-Weiner\thanks{equal contribution} \\ Brain and Cognitive Sciences \\ MIT \\ \ maxkw@mit.edu
  \And Michael L. Littman \\ Computer Science \\ Brown University \\ mlittman@cs.brown.edu
  \And Joshua B. Tenenbaum \\ Brain and Cognitive Sciences \\ MIT \\ jbt@mit.edu  }
\begin{document}

\maketitle

\begin{abstract}
  Human social behavior is structured by relationships. We form teams, groups, tribes, and alliances at all scales of human life. These structures guide multi-agent cooperation and competition, but when we observe others these underlying relationships are typically unobservable and hence must be inferred. Humans make these inferences intuitively and flexibly, often making rapid generalizations about the latent relationships that underlie behavior from just sparse and noisy observations. Rapid and accurate inferences are important for determining who to cooperate with, who to compete with, and how to cooperate in order to compete. Towards the goal of building machine-learning algorithms with human-like social intelligence, we develop a generative model of multi-agent action understanding based on a novel representation for these latent relationships called Composable Team Hierarchies (CTH). This representation is grounded in the formalism of stochastic games and multi-agent reinforcement learning. We use CTH as a target for Bayesian inference yielding a new algorithm for understanding behavior in groups that can both infer hidden relationships as well as predict future actions for multiple agents interacting together. Our algorithm rapidly recovers an underlying causal model of how agents relate in spatial stochastic games from just a few observations. The patterns of inference made by this algorithm closely correspond with human judgments and the algorithm makes the same rapid generalizations that people do. 
\end{abstract}

\section{Introduction}
Cooperation enables people to achieve together what no individual would be capable of on her own. From a group of hunters coordinating their movements to an ad-hoc team of programmers working on an open source project, the scale and scope of human cooperation behavior is unique in the natural world \cite{tomasello2014natural,henrich2015secret}. However, cooperation exists in a competitive world and finding the right balance between cooperation and competition is a fundamental challenge for anyone in a diverse multi-agent world. At the core of this challenge is figuring out who to cooperate with. How do we distinguish between friend and foe? How can we parse a multi-agent world into groups, tribes, and alliances? Typically when we observe behavior we only get information about these latent relationships sparsely and indirectly through the actions chosen by agents. Furthermore, these inferences are fundamentally challenging because of their inherent ambiguity; we are friend to some and foe to others \cite{galinsky2015friend}. They are also compositional and dynamic; we may cooperate with some agents in order to better compete against another. In order for socially aware AI systems to be capable of acting as our cooperative partners they must learn the latent structure that governs social interaction.

In some domains like sports and formal games, this social structure is known in advance and is essentially written into the environment itself e.g., ``the rules of the game'' \cite{kitano1997robocup,jaderberg2018human}. In contrast we focus on cases where cooperation is more ambiguous or could even be ad-hoc. In real life, people rarely play the same game twice and have to figure out the rules as they go along whether it's the ``rules of war'' or navigating office politics. 

Even young children navigate this uncertainty frequently and display spontaneous cooperation in novel situations from an early age \cite{warneken2006altruistic,hamlin2007social,hamann2011collaboration}. There is increasing evidence that this early arising ability to do social evaluation and inference relies on ``Theory-of-Mind'' (ToM) i.e., a generative model of other  agents with mental states of their own \cite{spelke2007core,kiley2013mentalistic}. People use these models to simulate what another agent might do next or consider what they themselves would do hypothetically in a new situation. From the perspective of building more socially sophisticated machines, ToM acts as a strong inductive bias for predicting actions. Rather than learning statistical patterns of low-level behavior which are often particular to a specific context (e.g., Bob often goes left, then up), an approach based on human ToM constrains inference to behaviors that are consistent with a higher-level mental state such as a goal, a belief or even a false-belief (e.g., Bob likes ice cream). When inference is carried out over these higher-level mental states, the inferences made are more likely to generalize to new contexts in a human-like way \cite{baker2009action,baker2017rational}.

Inspired by this ability, we aim to develop a new algorithm that applies these human-like inductive biases towards understanding groups of agents in mixed incentive (cooperative / competitive) contexts. These algorithms also serve as models of human cognition that can give us a deeper understanding of human social intelligence. Practically, by capturing the intuitive inferences that people make, our algorithm is more likely to integrate with humans since it will behave in predictable and understandable ways. Our approach builds on two major threads in the literature: generative models for action understanding and game theoretic models of recursive reasoning. Our contribution is the development of a new representation, Composable Team Hiearchies (CTH) for reasoning about how one agent's planning process depends on another and can flexibly capture the kinds of teams and alliances that structure group behavior. We propose that an algorithm using CTH as an inductive bias for Bayesian inverse planning  will have the flexibility to represent many types of group plans but is also constrained enough that it will enable the kinds of rapid generalizations that people do. We validate this hypothesis with two behavioral experiments where people are given the same scenarios as the algorithm and are asked to make the same inferences and predictions that the algorithm did. 

\subsection{Related Work}
Inferring the latent mental states of agents (e.g., beliefs, desires, and intentions) from behavior features prominently in machine learning and cognitive science (see \citet{albrecht2017autonomous} for a recent and comprehensive review from the machine learning point of view and \citet{jara2016naive} for a developmental perspective). Previous computational treatments similar in spirit to the approach here have focused on making inferences about other individuals acting in a single agent setting \cite{ng2000algorithms,baker2009action,ramirez2011goal,evans2016learning,nakahashi2016modeling,baker2017rational,rabinowitz2018machine}. When these tools are applied to multi-agent and game theoretic contexts they have focused on dyadic interactions \cite{yoshida2008game,ullman2009help,kleiman2016coordinate,raileanu2018modeling}. Dyadic interactions are significantly simpler from a representational perspective since an observer must merely determine whether each agent is cooperating or competing.

However, when the number of agents increases beyond a two player dyadic interaction, the problem of balancing cooperation and competition often takes on a qualitatively different character. Going from two to three or more players means the choice is no longer whether to simply cooperate or compete. Instead agents must reason about which agents they should cooperate with and which they should compete with.  In a dyadic interaction there is no possibility of cooperating \emph{to} compete or the creation of more complicated alliances and groups. 

\section{Computational Formalism}

\subsection{Stochastic Games}
We study multi-agent interactions in stochastic games which generalize single-agent  Markov Decision Processes to sequential decision making environments with multiple agents. Formally, a stochastic game, $G$, is the tuple $\tuple{n, \mathcal{S},\mathcal{A}_{1\ldots n}, T, R_{1\ldots n}, \gamma }$ where $n$ is the number of agents, $\mathcal{S}$ is a set of states, $\mathcal{A}_{1 \ldots n}$ is the joint action space with $\mathcal{A}_i$ the set of actions available to agent $i$, $T(s,a_{1 \ldots n},s')$ is the transition function which describes the probability of transitioning from state $s$ to $s'$ after $a_{1 \ldots n}$, $R_{1 \ldots n}(s,a_{1 \ldots n}, s')$ is the reward function for each agent, and $\gamma$ is the discount factor \cite{bowling2000analysis,filar2012competitive}. The behavior of each agent is defined by a policy $\pi_{1 \ldots n}(s)$ which is the probability distribution over actions that each agent will take in state $s$. 

There are many different notions of what it means to ``solve'' a stochastic game. Many of these concepts rely on notions of finding a best-response (Nash) equilibrium \cite{littman1994markov,littman2001friend,hu2003nash}. While solution concepts based on equilibrium analyses provide some constraints on the policies agents will use, they cannot provide a way to explain behavior carried out by bounded or cooperative agents who are willing to play a dominated strategy to help another agent. When games are repeated, there are often an explosion of equilibrium and these methods do not provide a clear method for choosing between them. Finally, there is ample evidence that both human behavior and judgments are not well explained by equilibrium thinking \cite{wright2010beyond}. On the other hand, without constraints from rational planning on the types of policies that agents are expected to use, there will be no way for an observer to generalize or predict how an agent's policy will adapt to a new situation or context that the agent has not been observed to act in. 

\begin{figure*}[tb]
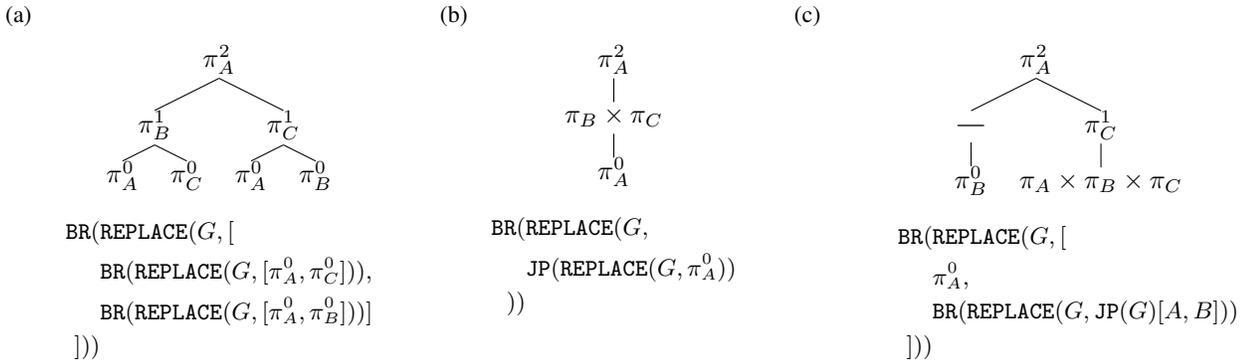

\centering
\captionsetup[sub]{justification=raggedright, singlelinecheck=false}
\begin{subfigure}[t]{.32\linewidth}
  \caption{\label{fig:treeK}}
    \Tree [.$\pi^2_A$ [.$\pi_B^1$ $\pi_A^0$ $\pi_C^0$ ] [.$\pi_C^1$ $\pi_A^0$ $\pi_B^0$ ] ]
    \\
    {\small
    \begin{align*}
    \BR(&\REP(G, [\\
    &\BR(\REP(G, [\pi^0_A, \pi^0_C])), \\
    &\BR(\REP(G, [\pi^0_A, \pi^0_B]))] \\
    ]))
    \end{align*}
    }
\end{subfigure}
\begin{subfigure}[t]{.26\linewidth}
    \caption{\label{fig:CoopComp}}  
    \Tree [.$\pi^2_A$ [.$\pi_B\times\pi_C$ $\pi_A^0$ ] ]
    \\
    {\small
    \begin{align*}
    \BR(&\REP(G, \\
    &\JP(\REP(G, \pi_A^0)) \\
    ))     
    \end{align*}
    }
\end{subfigure}
\begin{subfigure}[t]{.41\linewidth}
    \caption{\label{fig:Betray}}  
    \Tree [.$\pi^2_A$  [.| $\pi_B^0$ ] [.$\pi_C^1$ $\pi_A\times\pi_B\times\pi_C$ ] ]
    \\
    {\small
    \begin{align*}
    \BR(&\REP(G, [\\
    &\pi_A^0, \\
    &\BR(\REP(G, \JP(G)[A,B])) \\
    ]))     
    \end{align*}
    }
\end{subfigure}
    \caption{Example composition of base policies and operators to construct different types of teams with agents of variable sophistication. See text for descriptions of the operators and descriptions of these models. }
    \label{fig:trees}
\end{figure*}

\subsection{Group Plan Representation}
In this section we build up a representation for multi-agent interaction that can be used to compute policies for agents in novel situations, but is also sufficiently constrained that it can be used for rapid inference. We first introduce two simple planning operators based on individual best-response ($\BR$) and joint-planning ($\JP$). We then show how they can be composed together using a $\REP$ operator into Composable Team Hierarchies (CTH) which enable the flexible representation of teams and alliances within a MARL context. 

\subsubsection{Operator Composition ($\REP$)}
We first define the $\REP$ operator that takes an $N$ player stochastic game, $G$ and a policy $\pi_R$ indexed to a particular set of players ($R$) and returns an $N-|R|$ agent stochastic game $G'$ with the agents in $R$ removed. This new game $G'$ embeds $\pi_R$ in $G$ to generate dynamics for $R$ such that from the perspective of an agent in $G'$, the agents in $R$ are now stationary parts of the environment in $G'$ predictable from their policies. Formally $\REP(G, \pi_R)$ creates a game $G'$ identical to $G$ but with a reduced action space that excludes the agents in $R$ and modifies the transition function as follows:
\begin{align*}
  T_{G'}(s'|s, a_{-R}) =  \sum_{a_R} T_G(s'|s, a_{-R}, a_{R})\prod_{r \in R} \mathbbm{1}(\pi_r(s) = a_r)
\end{align*}
where the $-R$ refers to all agents other than those in $R$.

\subsubsection{Best Response ($\BR$)}
The best response operator $\BR$ takes a game $G$ with a single agent $i$ and returns a policy $\pi_i$ for just that agent. Planning here is defined through the standard Bellman equations. 
\begin{align*}
  &Q(s, a_i) = \sum_{s'} T(s'|s, a_i)[R_i(s, a_i, s') + \gamma \max_{a{_i}'} Q(s', a_i')]  \\
  &\pi_i(s) = \argmax_{a_i} Q(s, a_i)
\end{align*}
where ties are broken with uniform probability. 

\subsubsection{Joint Planning ($\JP$)}
A second planning operator $\JP$ generates cooperative behavior through joint planning. Each individual agent considers itself part of a hypothetical centralized ``team-agent'' that has joint control of all the agents that are included in the team and optimizes the joint reward of that team \cite{sugden2003logic,bratman2014shared}.  Planning under this approach combines all the agents in a game into a single agent and finds a joint plan which optimizes that group objective \cite{de2008polynomial,oliehoek2008optimal,kleiman2016coordinate}. If $J$ is the set of agents that have joined together as a team, their joint-plan can be characterized as:
\begin{align*}
  Q^J(s, a_J) = \sum_{s'} &T(s'|s, a_J) * \\
  &\Big(\sum_{j \in J}R_j(s, a_j, s') + \gamma \max_{a_{J}'} Q^J(s', a_{J}')\Big)
\end{align*}
Each agent plays its role in the team plan by marginalizing out the actions of all the other agents: $\pi_i(s) = \argmax_{a} Q^J(s, a)$
where ties are broken with uniform probability. Thus $\JP$ takes a $N>1$ agent game $G$ as input and returns policies $\pi_{J}$ as if all agents in $G$ are cooperating with each other towards a joint goal.

\subsubsection{Base Policies $\pi^0$}
Base policies ($\pi^0$) are non-strategic i.e., they take actions independent of the other agents. A common choice is to act randomly or to choose a locally optimal option ignoring all other agents. See \cite{wright2014level} for the various choices one could make about the level-0 policy in matrix-form games, some of which could be extended to stochastic games. 

\subsection{Composable Team Hierarchies}
We now show how with just these three simple operators ($\REP$, $\JP$, $\BR$) and a set of base policies ($\pi^0_{1 \ldots n}$) we can create complex team plans that vary in both their team structures as well as their sophistication. We start by noting that the output of $\JP$ and $\BR$ (policies) is an input to $\REP$, and the output of $\REP$ (games with fewer players) is an input to $\JP$ and $\BR$. When composed together, these operators generate hierarchies of policies.

Figure~\ref{fig:trees} shows how these planning procedures can be composed together to create strategic agents (using $\BR$), teams of cooperative agents (using $\JP$) and compositional combinations of the two. Even with just three players there are a combinatorial number of possible team partitions (all playing together, two against one, no teams) and higher and lower levels of sophisticated agents within those partitions. When shown hierarchically, this representation mirrors the tree-like structure of a grammar producing an ``infinite use of finite means'' -- the key benefit of a composable representation. We call these structures Composable Team Hierarchies (CTH). We now show how previous approaches from the literature can be subsumed under CTH, which unifies some previous approaches and enables new ways of reasoning about plans.

\subsubsection{Level-K Planning}
One technique common in behavioral game theory is iterative best response which is often called level-K or cognitive hierarchy \cite{camerer2004cognitive,wright2010beyond}. In brief, an agent operating at level $K$ assumes that other agents are using $K-1$ level policies.  This approach has also been extended to sequential decision making in reinforcement learning \cite{yoshida2008game,kleiman2016coordinate,lanctot2017unified}. By considering only a finite number ($K$) of these best responses, an infinite regress is prevented. This also captures some intuitive constraints on bounded thinking. This approach to multi-agent planning is to replace all other agents with a slightly less strategic $k-1$ policy. Importantly, this formalism maps a multi-agent planning problem into a hierarchy of nested single-agent planning problems. This recursive hierarchy grounds out in level-0 models ($\pi_i^0$) which we described above as base policies.

Figure~\ref{fig:treeK} shows how a level-K policy with $K=2$ for agent $A$ can be constructed by iterating between the $\BR$ and $\REP$ operators. The CTH shows a level-2 $A$ best responding to level-1 models of $B$ and $C$ who are best responding to level-0 base policies of $A$ \& $C$ and $A$ \& $B$ respectively. 

\subsubsection{Cooperative Planning}
While Level-K representations can capture certain aspects of strategic thinking i.e., how to best respond in one's own interest to other agents, it is not sufficient to generate the full range of social behavior. Specifically it will not generate cooperative behavior when cooperation is dominated in a particular scenario. However cooperative behavior between agents that form teams and alliances is commonly observed. An agent may be optimizing for a longer horizon where the game is repeated or one's reputation is at stake. Furthermore, certain agents may have intrinsic pro-social dispositions and an observer must be able to reason about these. A cooperative stance towards a problem can be modeled as a DEC-MDP \cite{oliehoek2008optimal}. In CTH this stance is easily represented. For instance starting with a base game $G$ that has players $(A,B,C)$ one can compute all three policies for working together as: $\JP(G)$.

\subsubsection{Composing Cooperation and Competition}
In addition to these two well studied formalisms, CTH can represent a range of possible social relationships that are not expressible with level-K planning or cooperative planning alone. Figure~\ref{fig:CoopComp} combines both operators to describe a cooperate \emph{to} compete stance. Under this CTH $A$ best responds to $B$ \& $C$ cooperating to compete against a naive version of $A$'s own behavior. Figure~\ref{fig:Betray} depicts agent $A$ best responding to both a naive $B$ and model of $C$ that is acting to betray the group of three. The CTH representation can capture an $A$ which is acting to counter a perceived betrayal by $C$. These examples show the representational flexibility of CTH and its ability to intuitively capture different social stances that agents might have towards each other. 

\subsection{Inverse Group Planning}
Observers can use CTH to probabilistically infer the various stances that each agent takes towards the others.  Agents represent their uncertainty over the CTH for agent $i$ as $P(\CTH_i)$, their prior beliefs before seeing any behavior. These beliefs can be updated in response to the observation of new behavioral data using Bayes rule:
\begin{align}
  P(\CTH_i | \mathbf{s}, \mathbf{a_i}) \propto  &P(\CTH)P(\mathbf{a_i} | \mathbf{s}, \CTH_i) \\
  = &P(\CTH_i) \prod_{t=1}^T P(a_{i,t}| s_t, \CTH_i)
\end{align}
where $\mathbf{s}$ and $\mathbf{a_i}$ are sequences of states and actions from time $1\ldots T$. $P(a_{i,t}| s_t |\CTH_i)$ is the probability of a given action under the induced hierarchy of goal-directed planning as determined by a given CTH. We use the Luce-choice decision rule to transform the Q-values of each action under planning into a probability distribution:
\begin{align}
    P(a_{i,t}| s_t, \CTH_i) \propto \exp(\beta * Q^*_\CTH(s,a))
\end{align}
where $\beta$ controls the degree to which the observer believes agents are able to correctly maximize their future expected utility at each time step. When $\beta \xrightarrow{} \infty$ the observer believes that agents are perfect optimizers, as $\beta \xrightarrow{} 0$ the observer believes the other agents are acting randomly. $Q^*_\CTH(s,a)$ are the optimal Q-values of the root agent in a given $\CTH$. 

In theory, the number of CTH considered by an observer could be infinite since the number of levels in the hierarchy does not have to be bounded. As this would make inference impossible, a realistic assumption is to assume some maximum level of sophistication which bounds the number of levels in the hierarchy \cite{yoshida2008game,kleiman2016coordinate}. Another possibility is to put a monotonically decreasing probability distribution on larger CTH as is done in the cognitive hierarchy model. Finally, since we have posed this IRL problem as probabilistic inference, Markov Chain Monte Carlo (MCMC) algorithms and other sampling approaches might enable the computation of $P(\CTH_i | \mathbf{s}, \mathbf{a_i})$ even when the number of hypothetical CTH are large. In this work we are agnostic to the how the polices are computed as any reinforcement learning algorithm is possible. In our simulations we used a simple version of Monte Carlo Tree Search (MCTS) based on Upper Confidence Bound applied to Trees (UCT) which selectively explores promising action sequences \cite{browne2012survey}.

\begin{figure*}[tb]
  \captionsetup[sub]{justification=raggedright, singlelinecheck=false, skip=-10pt}
  \newcommand{\gamesize}{.65}
  \newcommand{\plotsize}{.3}
  \newcommand{\hboost}{\hspace{1em}}
  \centering
  \begin{subfigure}[t]{.33\linewidth}
    \caption{}
    \begin{subfigure}[b]{\gamesize\linewidth}
      \includegraphics[page=1, trim={15cm 5cm 15cm 5cm}, clip, width = \linewidth]{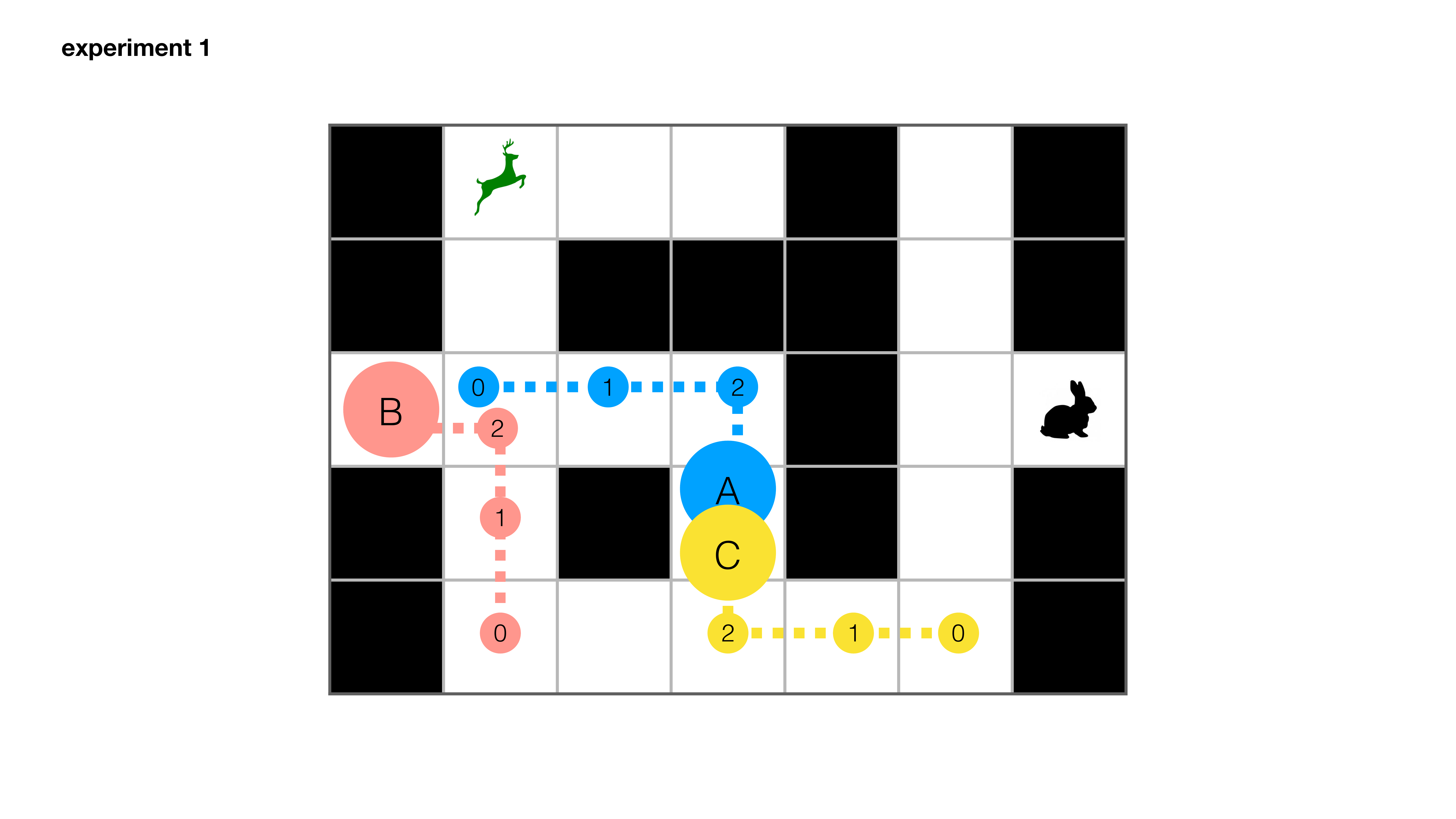}
      \hboost
    \end{subfigure}
    \begin{subfigure}[b]{\plotsize\linewidth}
      \includegraphics[width = \linewidth]{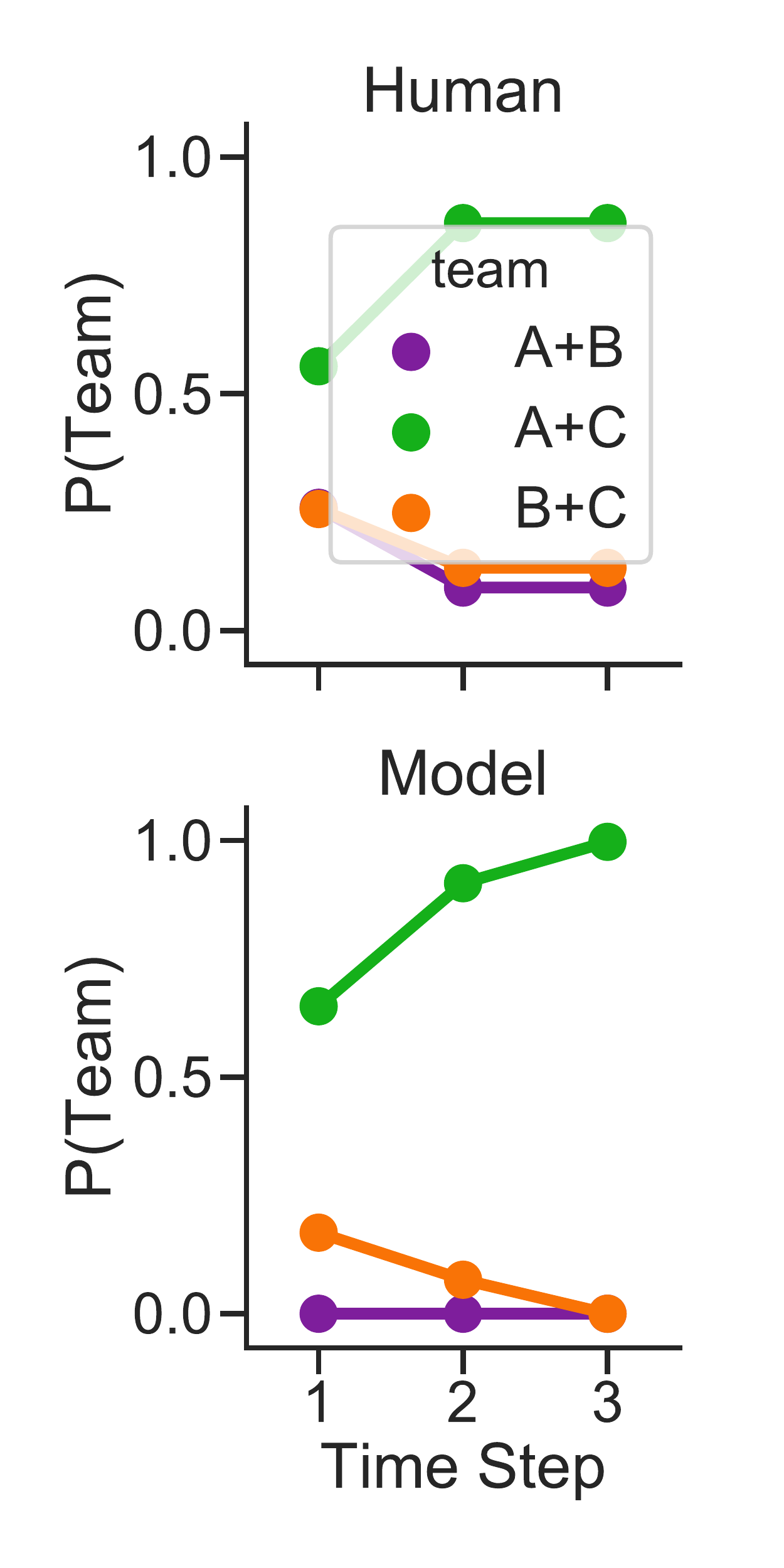}
    \end{subfigure}
  \end{subfigure}
  \begin{subfigure}[t]{.33\linewidth}
    \caption{}
    \begin{subfigure}[b]{\gamesize\linewidth}
      \includegraphics[page=2, trim={15cm 5cm 15cm 5cm}, clip, width = \linewidth]{figures_arxiv/grids}
      \hboost
    \end{subfigure}
    \begin{subfigure}[b]{\plotsize\linewidth}
      \includegraphics[width = \linewidth]{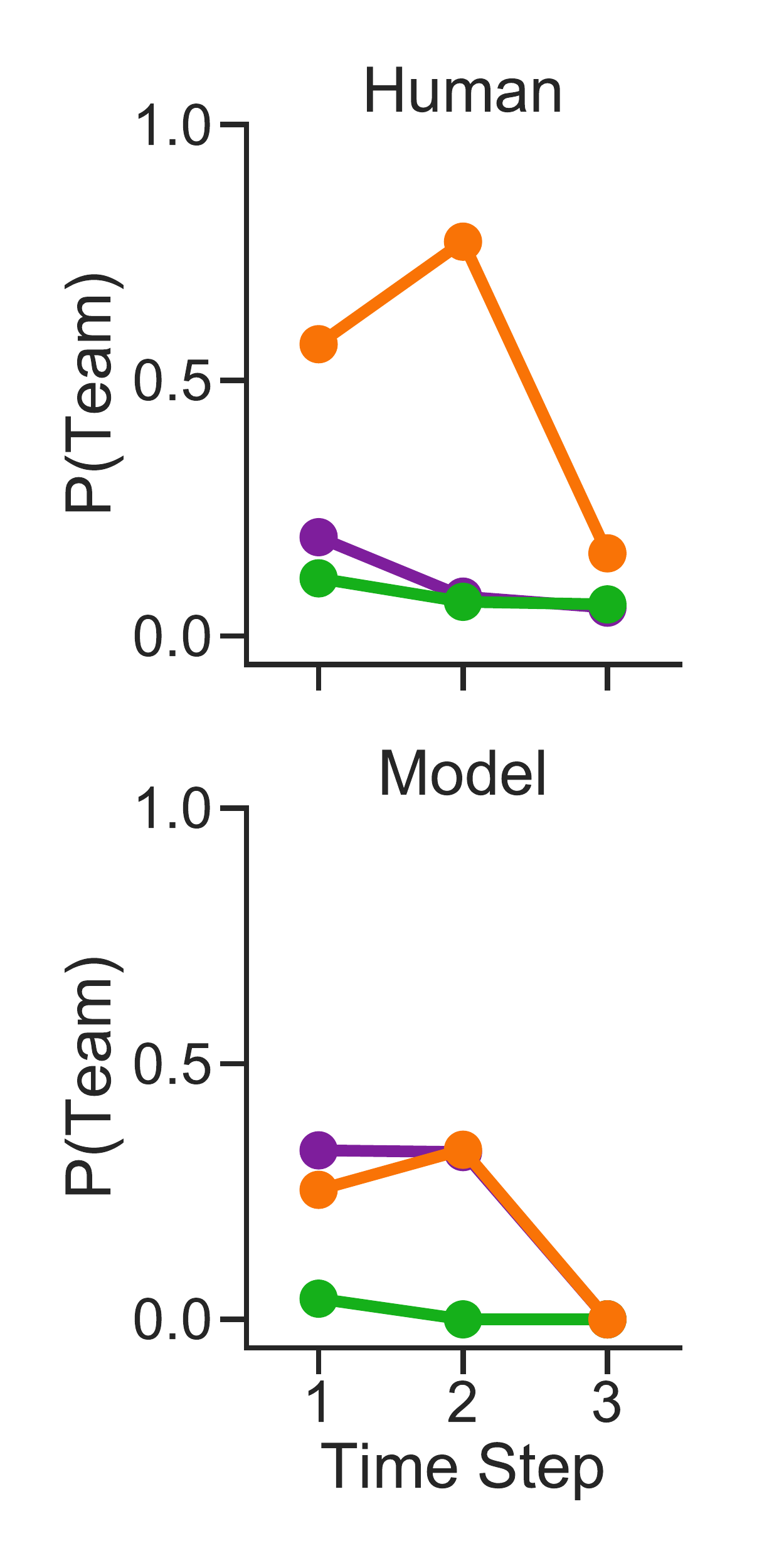}
    \end{subfigure}
  \end{subfigure}
  \begin{subfigure}[t]{.33\linewidth}
    \caption{}
    \begin{subfigure}[b]{\gamesize\linewidth}
      \includegraphics[page=3, trim={15cm 5cm 15cm 5cm}, clip, width = \linewidth]{figures_arxiv/grids}
      \hboost
    \end{subfigure}
    \begin{subfigure}[b]{\plotsize\linewidth}
      \includegraphics[width = \linewidth]{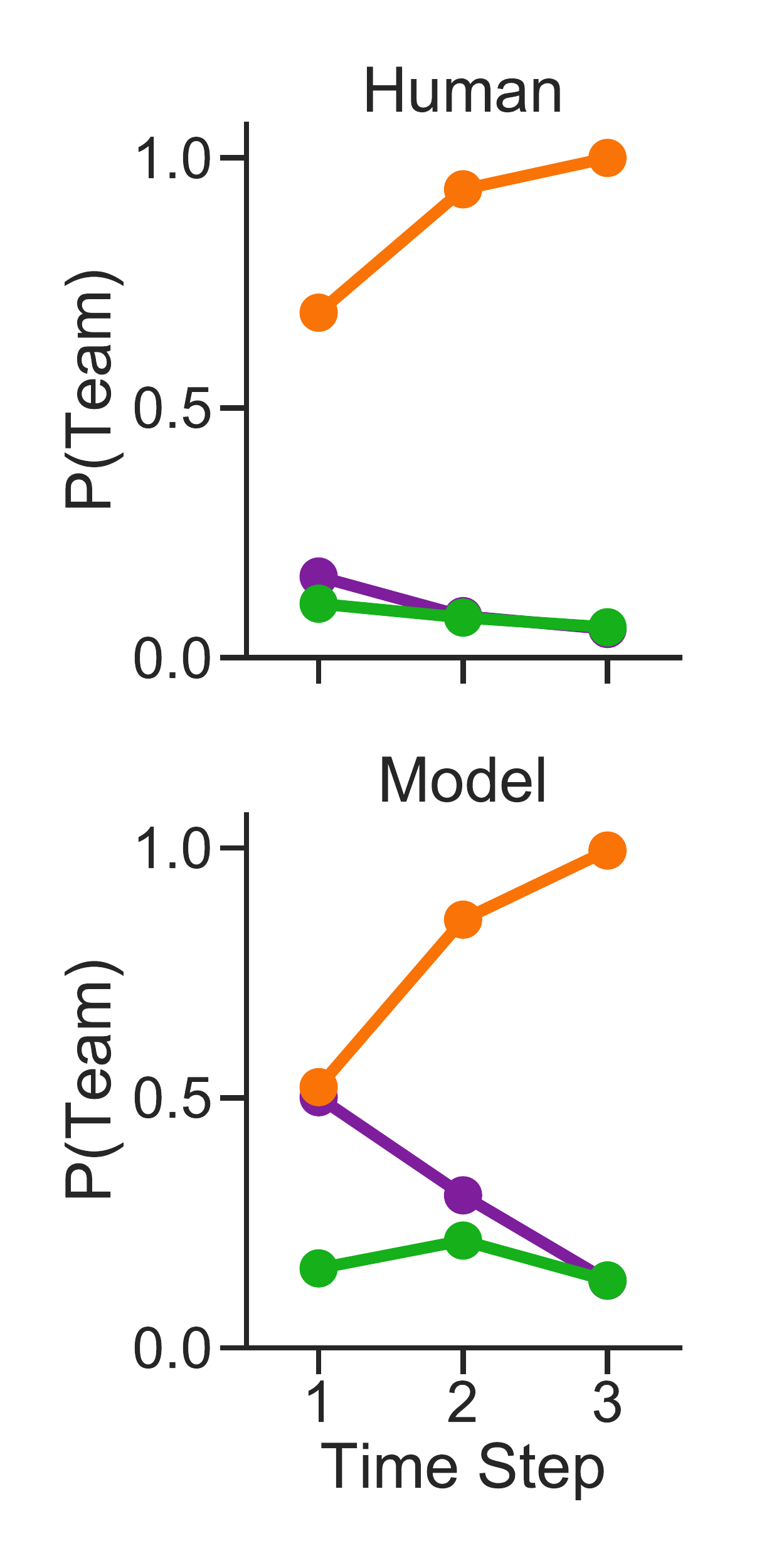}
    \end{subfigure}
  \end{subfigure}

  \begin{subfigure}[t]{.33\linewidth}
    \caption{}
    \begin{subfigure}[b]{\gamesize\linewidth}
      \includegraphics[page=4, trim={15cm 5cm 15cm 5cm}, clip, width = \linewidth]{figures_arxiv/grids}
      \hboost
    \end{subfigure}
    \begin{subfigure}[b]{\plotsize\linewidth}
      \includegraphics[width = \linewidth]{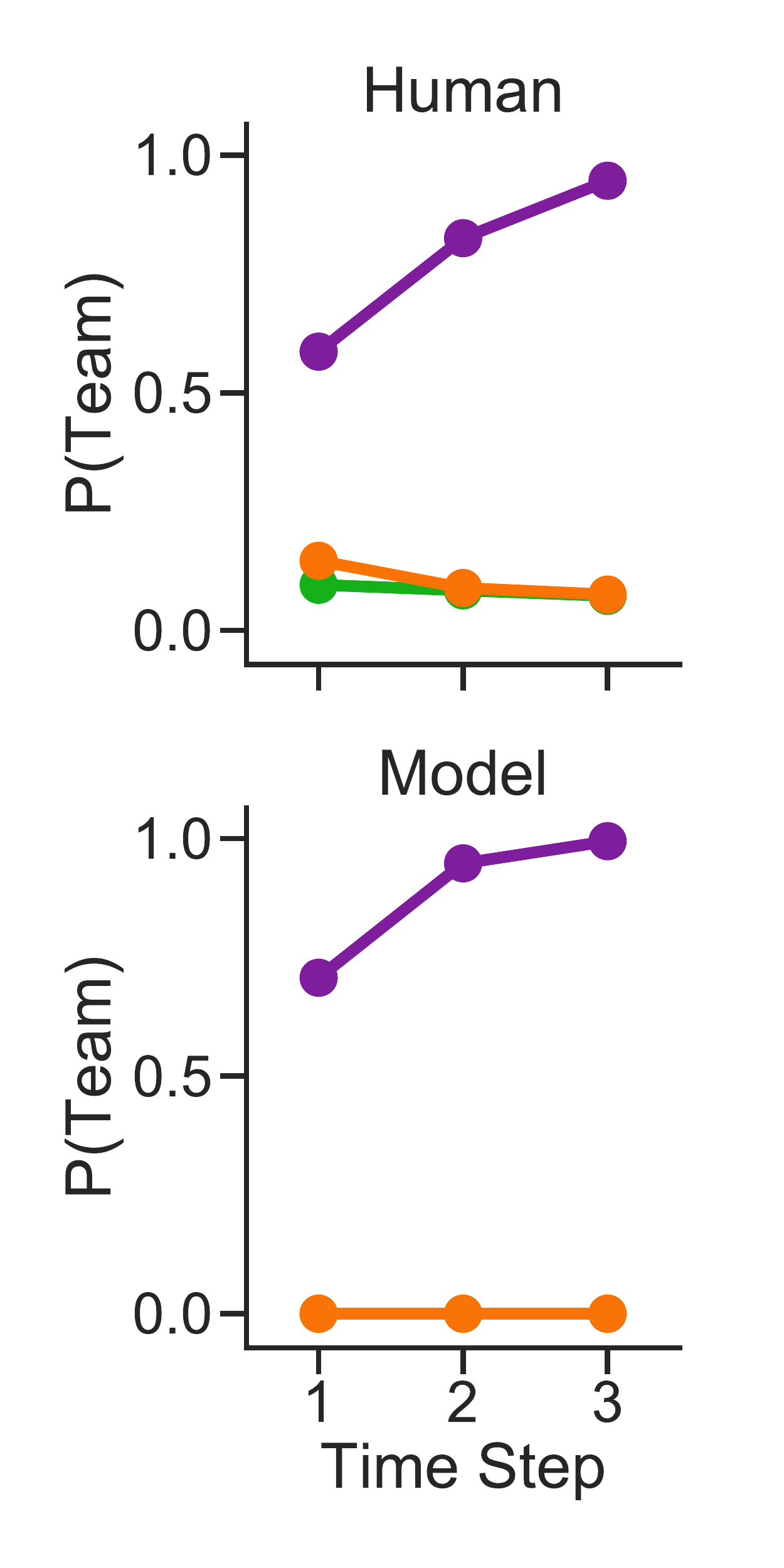}
    \end{subfigure}
  \end{subfigure}
  \begin{subfigure}[t]{.33\linewidth}
    \caption{}
    \begin{subfigure}[b]{\gamesize\linewidth}
      \includegraphics[page=5, trim={15cm 5cm 15cm 5cm}, clip, width = \linewidth]{figures_arxiv/grids}
      \hboost
    \end{subfigure}
    \begin{subfigure}[b]{\plotsize\linewidth}
      \includegraphics[width = \linewidth]{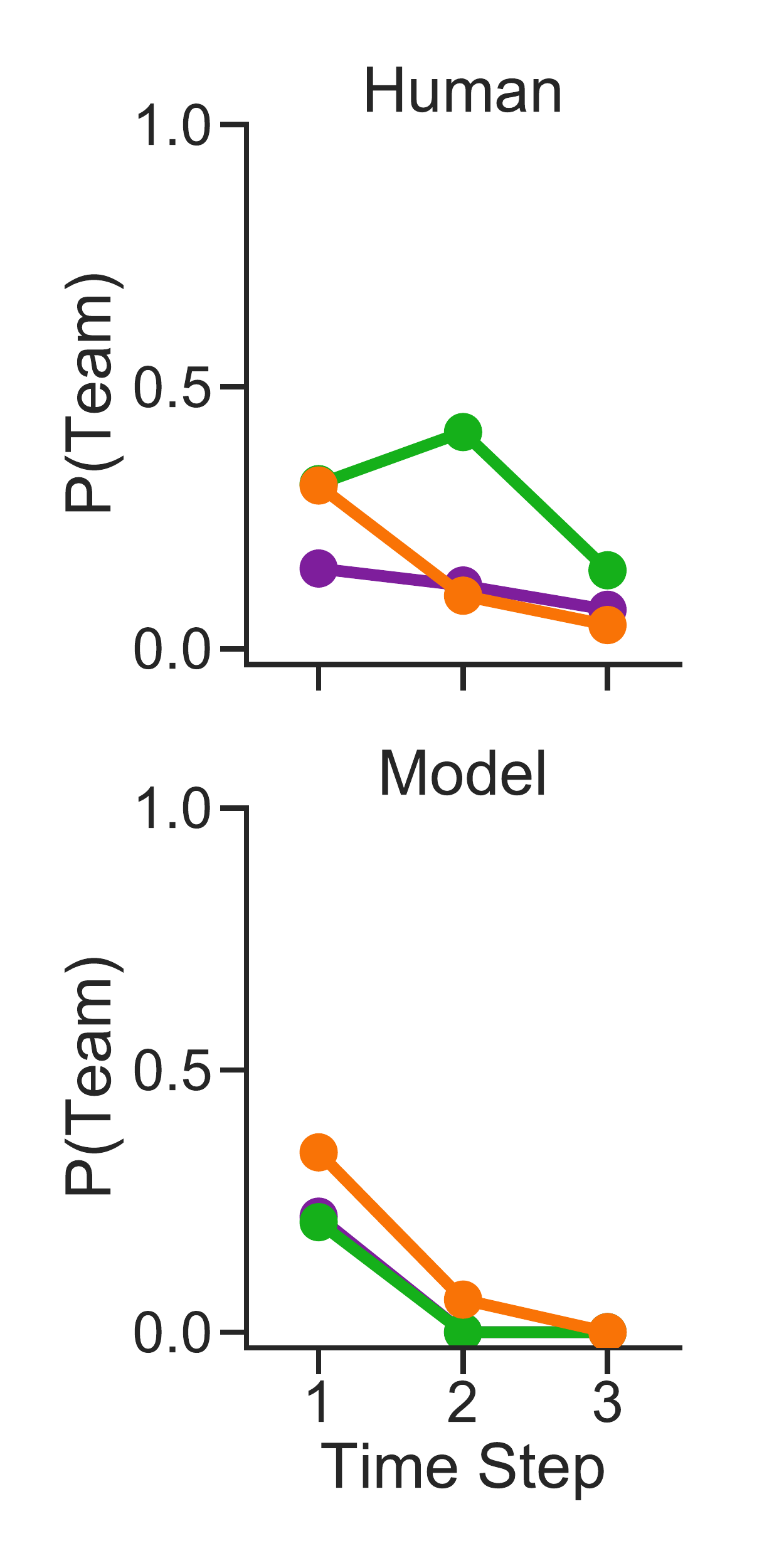}
    \end{subfigure}
  \end{subfigure}
  \begin{subfigure}[t]{.33\linewidth}
    \caption{}
    \begin{subfigure}[b]{\gamesize\linewidth}
      \includegraphics[page=6, trim={15cm 5cm 15cm 5cm}, clip, width = \linewidth]{figures_arxiv/grids}
      \hboost
    \end{subfigure}
    \begin{subfigure}[b]{\plotsize\linewidth}
      \includegraphics[width = \linewidth]{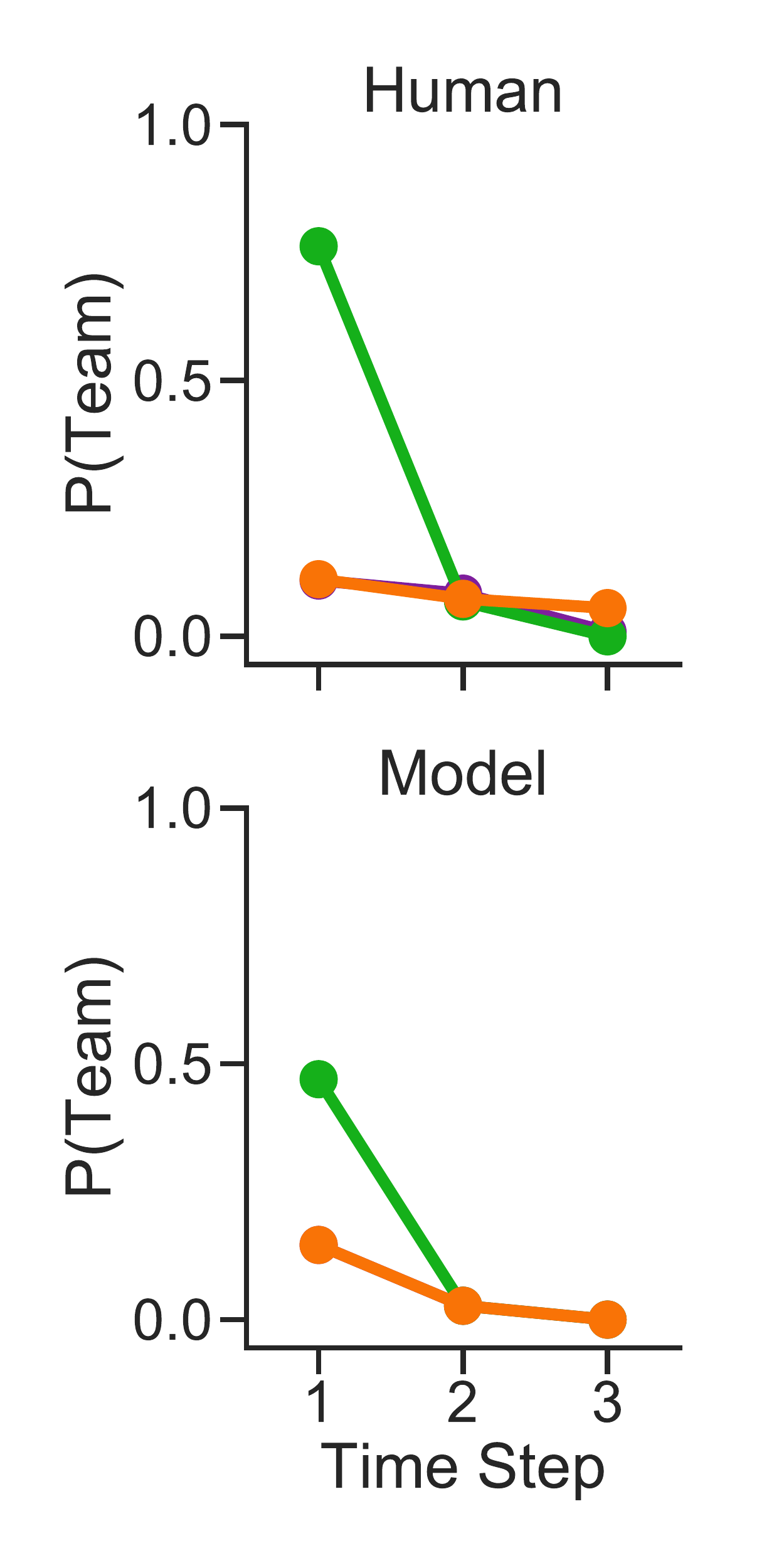}
    \end{subfigure}
  \end{subfigure}

    \begin{subfigure}[t]{.33\linewidth}
    \caption{}
    \begin{subfigure}[b]{\gamesize\linewidth}
      \includegraphics[page=7, trim={15cm 5cm 15cm 5cm}, clip, width = \linewidth]{figures_arxiv/grids}
      \hboost
    \end{subfigure}
    \begin{subfigure}[b]{\plotsize\linewidth}
      \includegraphics[width = \linewidth]{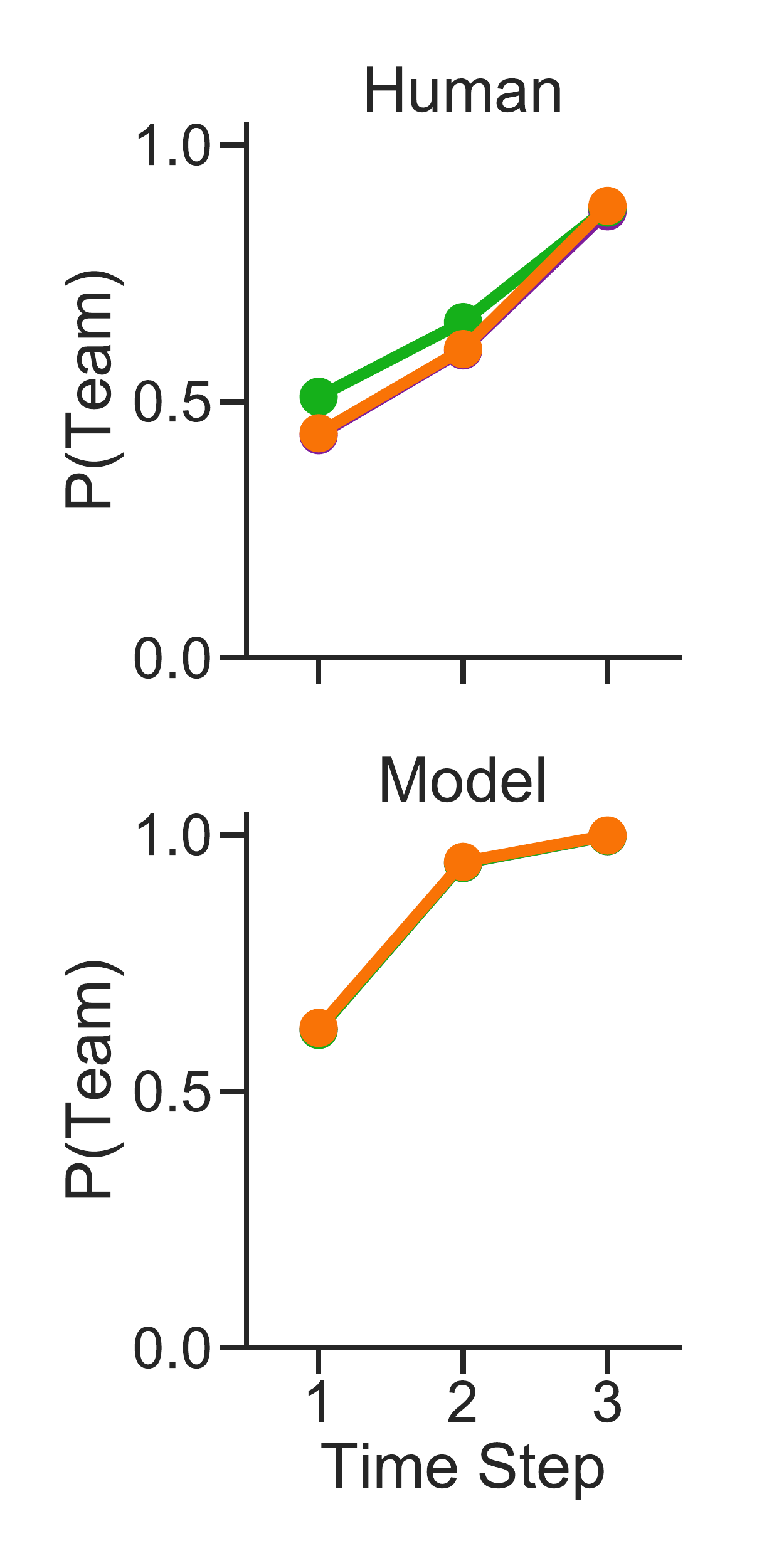}
    \end{subfigure}
  \end{subfigure}
  \begin{subfigure}[t]{.33\linewidth}
    \caption{}
    \begin{subfigure}[b]{\gamesize\linewidth}
      \includegraphics[page=8, trim={15cm 5cm 15cm 5cm}, clip, width = \linewidth]{figures_arxiv/grids}
      \hboost
    \end{subfigure}
    \begin{subfigure}[b]{\plotsize\linewidth}
      \includegraphics[width = \linewidth]{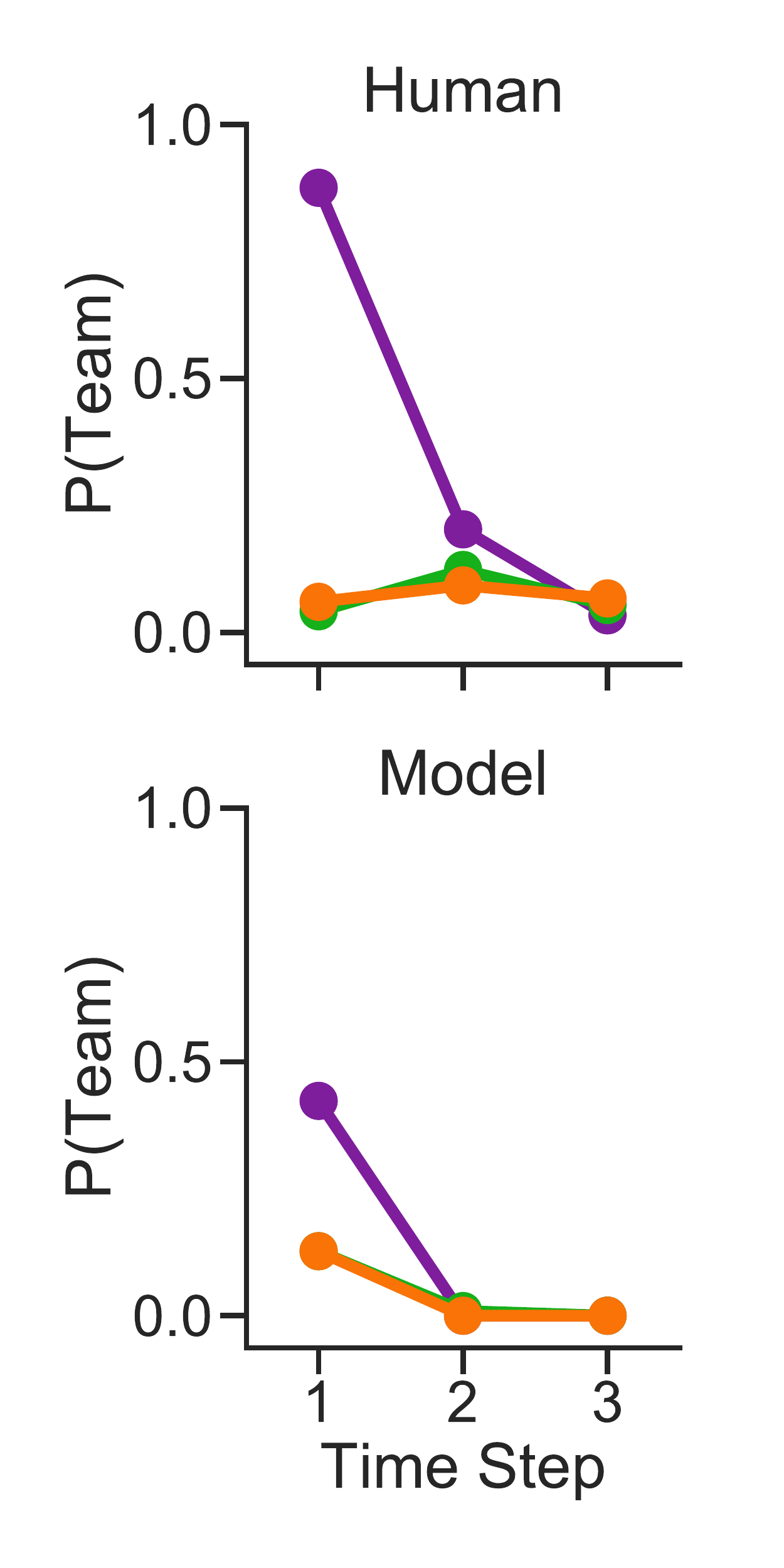}
    \end{subfigure}
  \end{subfigure}
  \begin{subfigure}[t]{.33\linewidth}
    \caption{}
    \begin{subfigure}[b]{\gamesize\linewidth}
      \includegraphics[page=9, trim={15cm 5cm 15cm 5cm}, clip, width = \linewidth]{figures_arxiv/grids}
      \hboost
    \end{subfigure}
    \begin{subfigure}[b]{\plotsize\linewidth}
      \includegraphics[width = \linewidth]{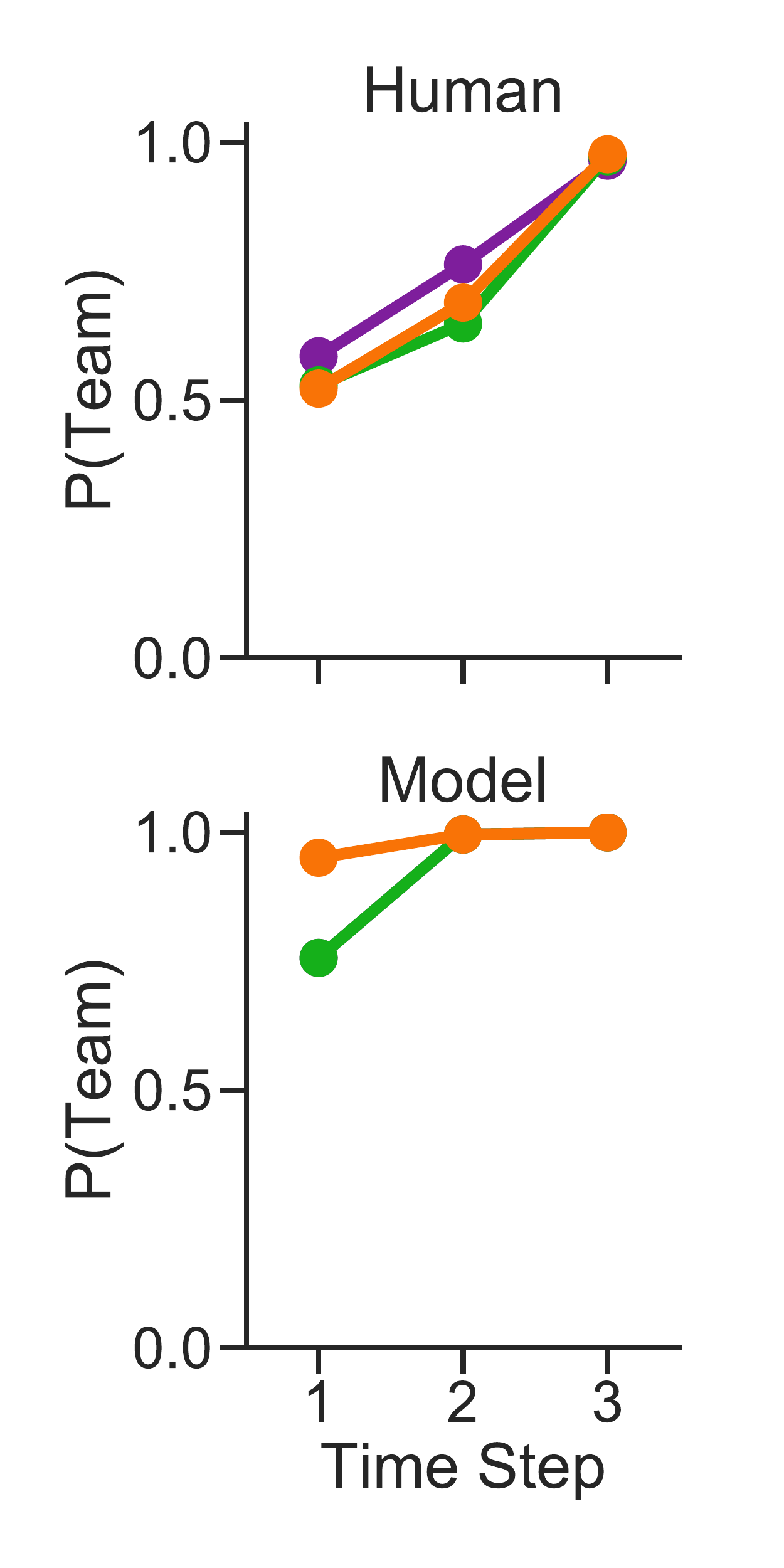}
    \end{subfigure}
  \end{subfigure}
  \caption{Experimental scenarios and results for Experiment 1. Each scene involved 3 time steps from a starting state. The hunters were represented with circles. Each agent's movement path is traced out with dotted lines with small numbered dots indicating the position in a given time step. Human participants saw a more dynamic scene that played out over the course of the experiment instead of seeing the full trajectory at once. To the right of each scenario are plots showing the average human inferences (top) and Bayesian model average results run on the same scene that participants saw. \label{fig:exp1} }
\end{figure*} 

\section{Experiments}
Our two experiments were carried out in a spatial stag-hunt domain. In these scenarios, the agents control hunters who gain rewards by capturing prey: stags and hares. The hunters and stags can move in the four cardinal directions and can stay in place. The hares do not move but the stags move away from any nearby hunters so must be cornered to be captured. All moves happen simultaneously and agents are allowed to occupy the same squares. 
 
Hunters are rewarded for catching either type of prey by moving into its square, and when any prey is caught the game terminates. Catching a hare is worth one point to the hunter who catches it. Catching a stag requires coordination but hunters split 20 points when capturing it. At least two hunters must simultaneously enter the square with the stag on it to earn the points. The stag-hunt is a common example used to demonstrate the challenges of coordinating on joint action and the creation of cooperative cultural norms \cite{skyrms2004stag}. Previous work on the spatial stag-hunt has mostly focused on modeling behavior, not inferences, in two-player versions. As noted before, with only two hunters there are limited possibilities for different team arrangements \cite{yoshida2008game,peysakhovich2018prosocial}.

We designed nine different scenes in this stag-hunt world that each had three hunters, two stags, and two hares (Figure~\ref{fig:exp1}). Different scenes had different starting positions and different spatial layouts which means that both people and our algorithm must generalize across different environments and contexts. This makes it less likely that a heuristic approach based on features will work. Instead we can test whether people invert an underlying group planning process. In both experiments (N=37, all USA), each participant was tested on all nine scenarios. Participants watched the scene unfold a few moves at a time. All human data was collected on Amazon Mechanical Turk. In Experiment 1 we compare our algorithm against the inferences people made about the underlying structure i.e., who is cooperating with who. In Experiment 2 we compare our algorithm against people's ability to predict the next action in a scene. In both experiments we compare human participant data with our Bayesian inference with a uniform prior over depth-1 CTH. When modeling human judgments and behavior $\beta$ usually corresponds to the noise in the utility maximization process where non-optimal decisions are made proportional to how close their utility is to optimal. In the team inference experiment (experiment 1) participants used a continuous slider to report their judgments so $\beta=1$ was used for model comparison while in the action-prediction experiment (experiment 2) subjects made a discrete choice to report their predictions so a higher $\beta=5$ was used. 

\subsection{Experiment 1: Team Inference}
For each scenario, participants made judgments about whether A\&B, B\&C, and C\&A were cooperating at three different time points by selecting a range between 0 and 100 on a slider. These ratings were averaged together and normalized to 0 and 1. In total, we collected a total of 81 distinct data points. Figure~\ref{fig:exp1} shows the human data and model results for each of the nine scenarios at each time point. We did not find any systematic individual differences among human participants. 

The model performs well across all inferred team structures: when all three players are on the same team (g, i), when just two are working together (a, c, d) and when all three are working independently (b, e, f, h). These inferences were made based on information about the abstract team structure since they were made before any of the actual outcomes were realized. Even a single move was often sufficient to give away the collective intentions of the group of hunters. Finally, the model also handles interesting reversals. In situation (b), one might infer that B and C were going to corner the stag but this inference is quickly overturned when C goes for a hare instead at the last minute. Situation (h) also contains a change of mind. At first it seems A is following B to capture the stag together but he then reverses and goes for the hare. Just this single reversal was sufficient to flip people's judgments about the underlying team structure and our algorithm captures this.

\begin{figure}[tb]
  \centering
  \captionsetup[sub]{justification=raggedright, singlelinecheck=false,  skip=-1pt}
  \begin{subfigure}[b]{.49\linewidth}
    \caption{\label{fig:BMA-CTH-TEAM}}
    \includegraphics[width=\linewidth]{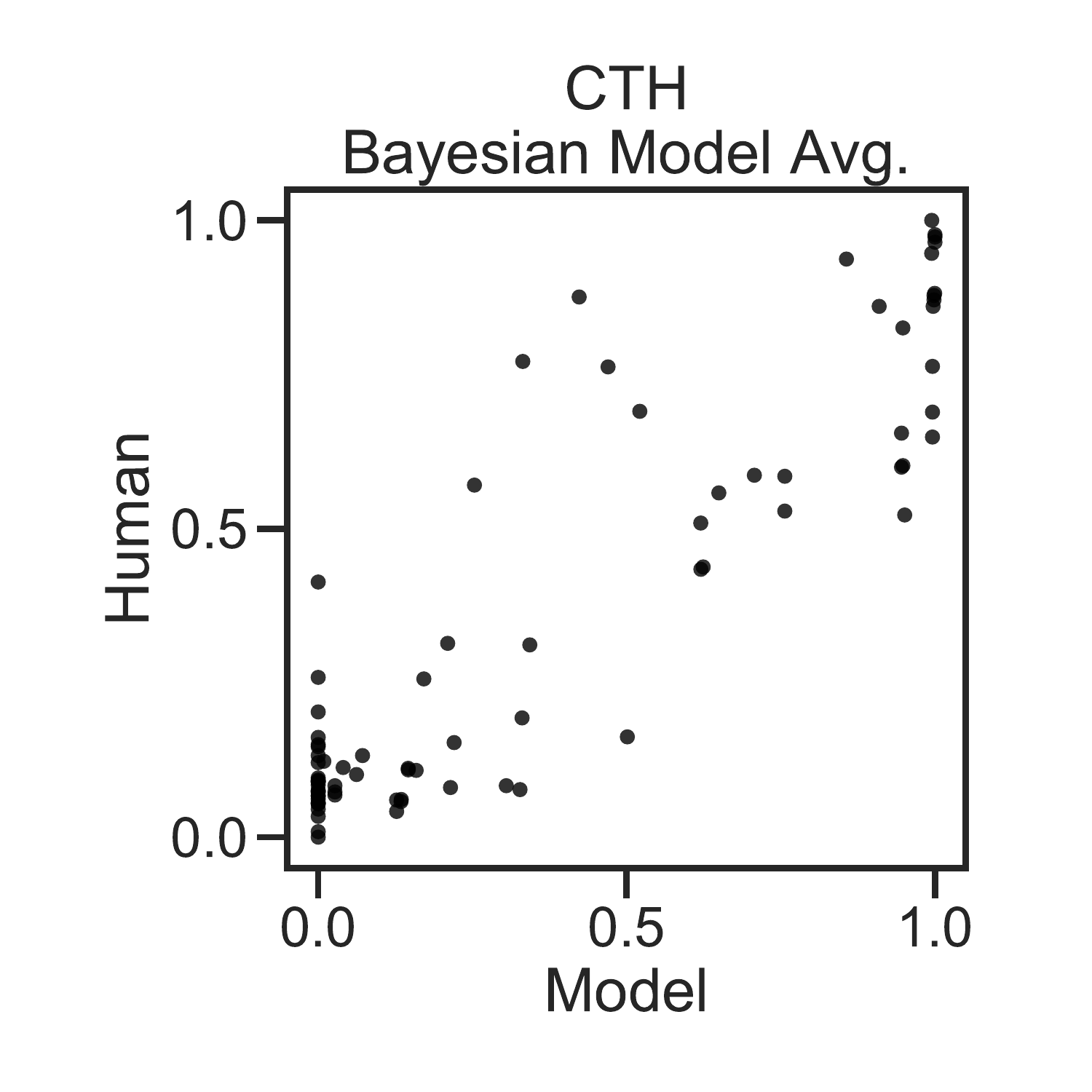}
    \end{subfigure}
    \begin{subfigure}[b]{.49\linewidth}
      \caption{\label{fig:BEST-CTH-TEAM}}
  \includegraphics[width=\linewidth]{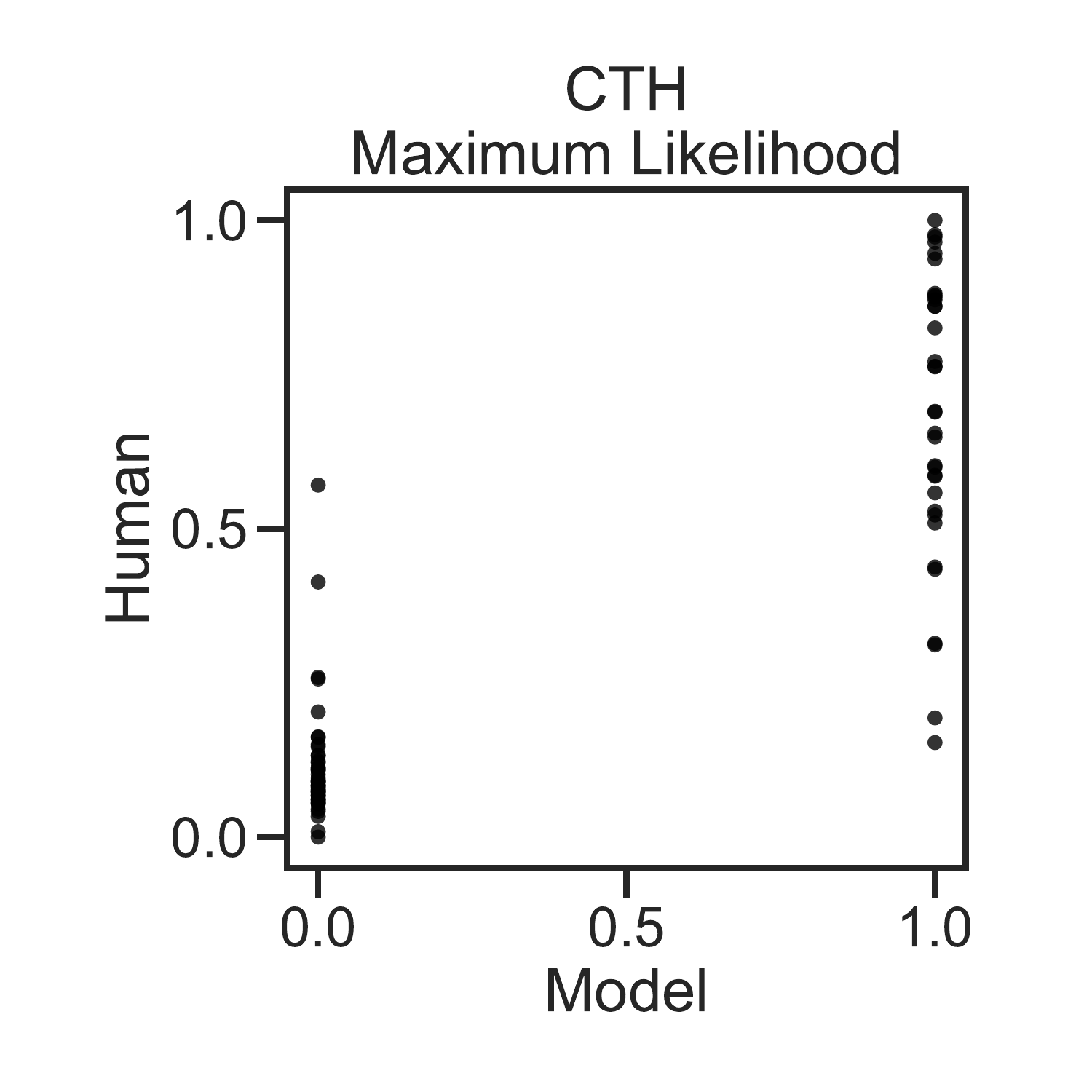}
    \end{subfigure}
  \caption{Quantification of algorithm inferences in Experiment 1 compared to human judgments. (a) Bayesian model averaging explains a high degree of variance in human judgments. (b) The maximum likelihood CTH captures some of the coarse grained aspects of the inferences but does not capture the uncertainty in people's judgments. \label{fig:team_scatter} }
\end{figure}

There were also a few circumstances where people's judgments significantly deviated from our algorithm.
For instance, in scenario (c) people were quicker to infer that B and C are on the same team while the model has greater uncertainty. This difference might reflect the fact that people put a higher prior on cooperative CTH while we used a uniform prior. Indeed, increasing the prior on cooperative CTH results in more human-like inferences for this scenario. Most of the differences were more subtle. The model makes stronger predictions (closer to 0 and 1) while people integrated information more slowly and less confidently. Figure~\ref{fig:team_scatter} and Table~\ref{tab:R1} show a quantification of how well our algorithm can act as a model for human judgments. Bayesian model averaging across the uncertainty in the underlying CTH (Figure~\ref{fig:BMA-CTH-TEAM}) did a better job of capturing human team inferences than did the CTH with the highest likelihood (Figure~\ref{fig:BEST-CTH-TEAM}). Thus a full Bayesian approach seems to be needed to capture the nuanced and graded judgments that people make as they integrate over the ambiguous behavior of the agents.

\subsection{Experiment 2: Action Prediction}
In a second experiment using the same set of stimuli as Experiment 1 we compared our system's ability to predict the next action with people's predictions. Each participant was given the choice to select the action (from those available) for each of the 3 hunters. Averaging over the participants gives a distribution over the next action for human participants. Across all nine scenarios, we elicited 53 judgements which generated 216 distinct data points. Since our computational formalism is a generative model the same algorithms that was used for team inferences is also tested on action predictions.

\begin{figure}[tb]
  \centering
  \captionsetup[sub]{justification=raggedright, singlelinecheck=false,  skip=-1pt}
  \begin{subfigure}[b]{.32\linewidth}
    \caption{\label{fig:BMA-ACTION}}
    \includegraphics[width=\linewidth]{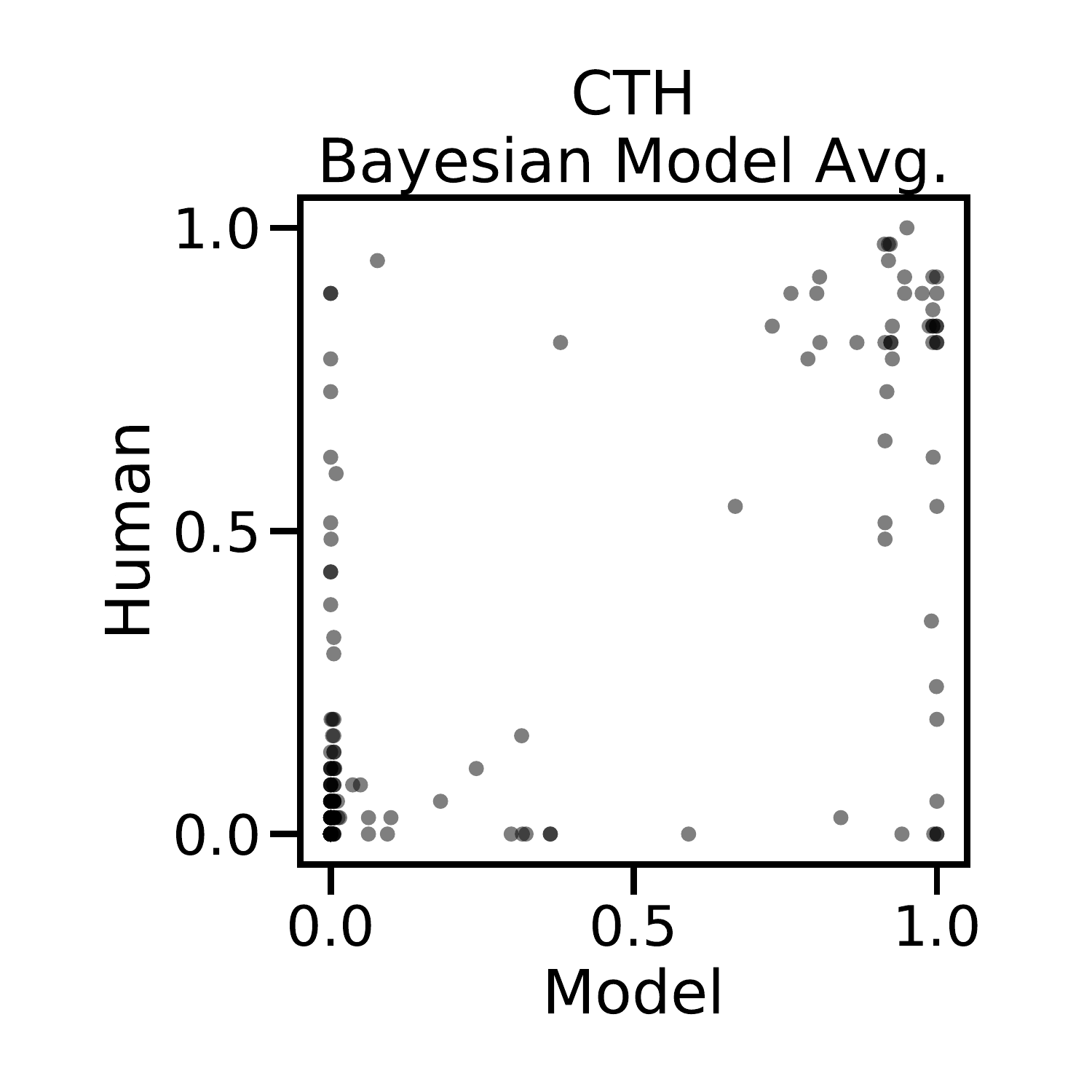}
  \end{subfigure}
  \begin{subfigure}[b]{.32\linewidth}
    \caption{\label{fig:BEST-ACTION}}
    \includegraphics[width=\linewidth]{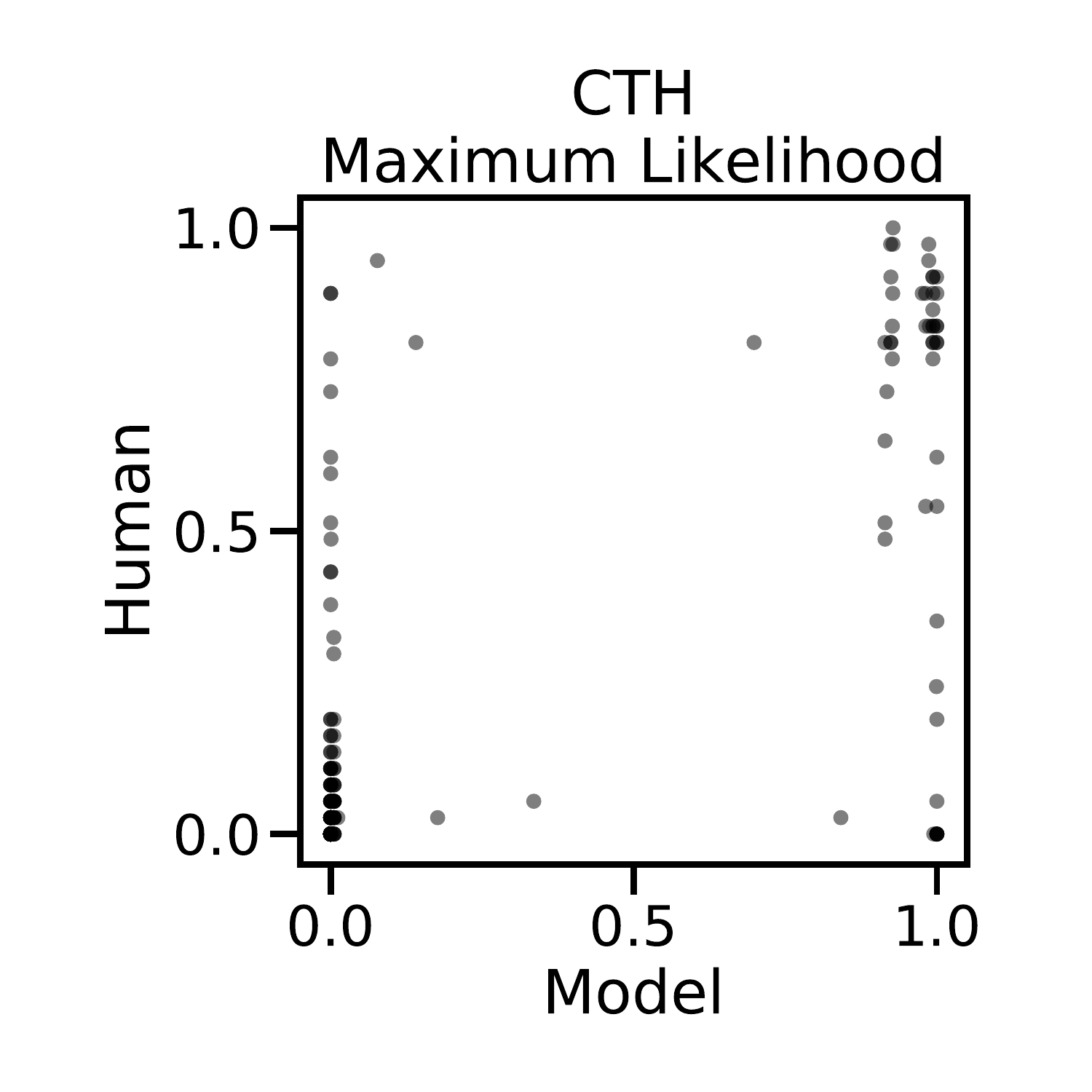}
  \end{subfigure}
  \begin{subfigure}[b]{.32\linewidth}
    \caption{\label{fig:NULL-ACTION}}
    \includegraphics[width=\linewidth]{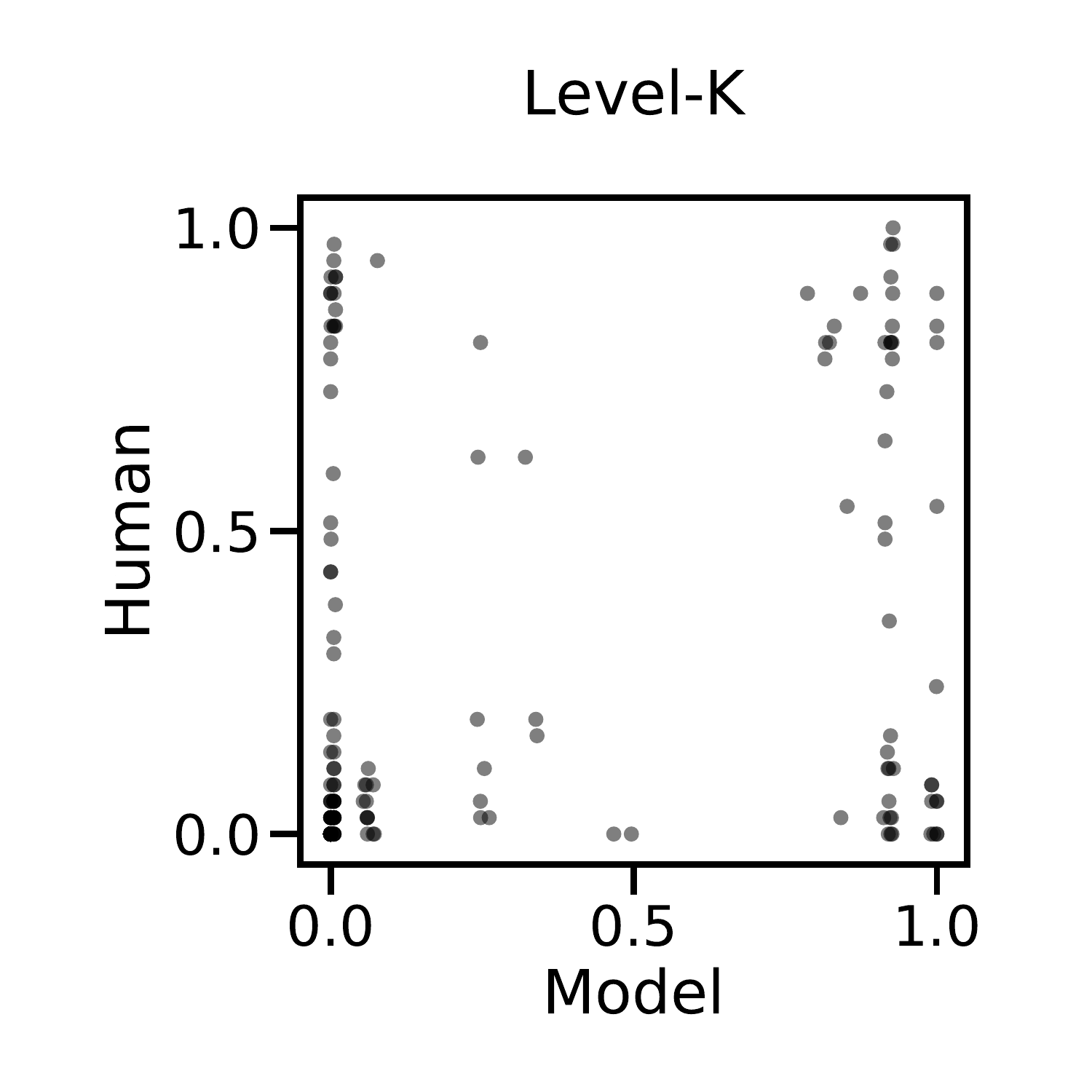}
  \end{subfigure}

  \caption{Quantification of algorithm predictions in Experiment 2 compared to human predictions. Both (a) Bayesian model averaging and (b) the maximum likelihood CTH explains a high degree of variance in human predictions. (c) Level-K models explain significantly less variance in human predictions.  \label{fig:action_scatter} }
\end{figure}

Figure~\ref{fig:action_scatter} and Table~\ref{tab:R2} show the ability of this algorithm to predict the human action prediction distribution. We find a relatively high-correlation ($R>0.7$) but we do not find as large of a difference between the Bayesian model averaging over CTH and the maximum likelihood CTH. This is likely due to the fact that each participant directly selected the action they thought was most likely rather than give a graded measure of confidence. Still both of these CTH based models out-perform a Level-K model which does not allow for the use of the $\JP$ operator. 

  \captionsetup[sub]{skip=5pt}
\begin{table}
  \centering
\begin{subtable}{\linewidth}\centering
\caption{Experiment 1\label{tab:R1}}
\begin{tabular}{r|cc}
\ & BMA-CTH & MLE-CTH \\
\hline R & \textbf{0.90} & 0.86 \\
RMSE & \textbf{0.18} & 0.28 \\
\end{tabular}
\end{subtable}

\begin{subtable}{\linewidth}\centering
\caption{Experiment 2 \label{tab:R2}}
\begin{tabular}{r|ccc}
\ & BMA-CTH & ML-CTH & Level-K\\
\hline R & \textbf{0.74} & 0.73 & 0.38 \\
RMSE & \textbf{0.27} & 0.28 & 0.41 \\
\end{tabular}
\end{subtable}
\caption{Pearson correlation coefficients (R) and root mean square error (RMSE) for both experiments. Higher R and lower RMSE indicate explaining more of the variation in the human (a) judgments and (b) predictions. BMA is Bayesian model average and ML is maximum likelihood. \label{tab:R}}
\end{table}

\section{Discussion}
Our contribution is a novel representation for extending single-agent generative models of action understanding to the richness of multi-agent interaction. In human cognition, the ability to infer the mental states of others is often called Theory-of-Mind and here we develop a Theory-of-Minds which explains and predicts group behavior. The core of this work is to build on two key insights from how people learn in general and in particular learn about other people \cite{tenenbaum2011grow}:
\begin{enumerate}
\item Agents have the ability to construct generative models of other agents and use those models to reason about future actions by simulating their planning. With models of other agents and a model of the environment, agents can predict what will happen next through forward simulation. With these future-oriented predictions of what other agents will do, individuals can better generate their own plans. They can even use these models hypothetically in order to predict what an agent would do, or counterfactually to predict what another agent would have done. 
\item The inferences people make about others take place at a high level of abstraction. For instance, people learn about who is cooperating with whom and why, rather than reasoning directly about the likelihood of a particular sequence of actions in a specific situation. While in some sense, these abstract inference are more complex, they drastically reduce a hypotheses space about every action an agent might take to a much smaller hypothesis space of actions that serve a social purpose. These abstract reasons generalize in ways that mere patterns of behavior do not. 
\end{enumerate}
In this work, we took inspiration from the ways that human observers think abstractly about alliances, friendships, and groups. We formalized these concepts in a multi-agent reinforcement learning formalism and used them as priors to make groups tractably understandable to an observer. Our model explains the fine-grained structure of human judgment and closely matches the predictions made by human observers in a novel and varying three-agent task. 

There are still many avenues for future work. While the approach described here can work well for small groups of agents, the computations involved scale poorly with the number of agents. Indeed, when interacting with a large number of agents more coarse-grained methods which ignore individual mental states might be required \cite{yang2018deep}. Another way forward is to constrain the possible types of CTH to consider. For instance, when dealing with a large number of agents, people seem to use group membership cues, some of which are directly observable such as style of dress or easily inferable such as language spoken \cite{liberman2017origins}. These cues could rapidly prune the number of CTHs considered but also could lead to biases. Another possible route to scaling these methods is through sophisticated state abstractions such as those in deep multi-agent reinforcement learning where agents are trained for cooperation and competition \cite{leibo2017multi,perolat2017multi,lowe2017multi,lerer2017maintaining,peysakhovich2018prosocial,foerster2018learning}. Self-play and feature learning based methods might also be useful for generating interesting base policies to build on in our CTH representation \cite{hartford2016deep}.

Our current set of experiments looked at situations where team coordination required spatial convergence. Future work will look at environments with buttons that can open and close doors or by giving agents the ability to physically block others. In these scenarios heuristics based on spatial convergence will not correlate with human judgments and higher order CTH may be needed to identify the underlying team structures. Finally, endowing multi-agent reinforcement learning agents with the ability to do CTH inference could give these agents the ability to more effectively reason about and coordinate with others. 

While we propose a method of understanding and planning with agents that have known teams, agents are frequently in scenarios where team structures have yet to be established e.g., the first day of kindergarten. In future work, we hope to explore how agents can identify the best teammates in an environment and create a social relationship with them. Based on previous actions an observer could begin to predict an agent's likelihood of changing its social stance through changes in the CTH structure. Finally, social norms and anti-social behavior such as punishing and disliking are not easily captured in the current version of the CTH representation. Future work will extend CTH with new operators that expand its flexibility. 

\subsubsection{Acknowledgments}
This work was supported by a Hertz Foundation Fellowship, the Center for Brains, Minds and Machines (CBMM), NSF 1637614, and DARPA Ground Truth. We thank Mark Ho for comments on the manuscript and for helping to develop these ideas. 

\fontsize{9pt}{10pt}\selectfont 
\bibliography{aaai19.bib}
\bibliographystyle{aaai}

\end{document}